\documentclass[runningheads]{llncs}
\usepackage{graphicx}

\usepackage{tikz}
\usepackage{comment}
\usepackage{amsmath,amssymb} 
\usepackage{color,colortbl}
\usepackage{pifont}
\usepackage{xcolor}
\usepackage{subfig}

\usepackage[accsupp]{axessibility}  

\newcommand{\cmark}{\ding{51}}%
\newcommand{\xmark}{\ding{55}}%

\definecolor{fyxcolor}{RGB}{0,128,255}
\definecolor{demphcolor}{RGB}{100,100,100}
\definecolor{citecolor}{RGB}{0,0,192} 
\definecolor{GrayBG}{gray}{0.90}

\def\eg{e.g.}
\def\ie{i.e.}
\def\etc{etc.}
\def\vs{vs.}
\def\x{$\times$}
\newcommand{\demph}[1]{\textcolor{demphcolor}{#1}}
\newcommand{\app}{\raise.17ex\hbox{$\scriptstyle\sim$}}
\newlength\savewidth\newcommand\shline{\noalign{\global\savewidth\arrayrulewidth
  \global\arrayrulewidth 1pt}\hline\noalign{\global\arrayrulewidth\savewidth}}
\newcommand{\tablestyle}[2]{\setlength{\tabcolsep}{#1}\renewcommand{\arraystretch}{#2}\centering\footnotesize}

\usepackage[pagebackref=true,breaklinks=true,letterpaper=true,colorlinks,hidelinks,bookmarks=false]{hyperref}
\usepackage{orcidlink}

\begin{document}

\newcommand{\ours}{MaCLR}
\newcommand{\rev}[1]{{\color{red}#1}}
\iftrue
\newcommand{\fy}[1]{{\color{red}[FY: #1]}}
\newcommand{\dvd}[1]{\textcolor{blue}{\bf [DAVIDE: #1]}}
\newcommand{\todo}[1]{\textcolor{green}{\bf [TODO: #1]}}
\newcommand{\joe}[1]{\textcolor{magenta}{[Joe: #1]}}
\else
\fi

\pagestyle{headings}
\mainmatter
\def\ECCVSubNumber{6218}  

\title{\ours{}: Motion-aware Contrastive Learning of Representations for Videos} 

\titlerunning{\ours{}: Motion-aware Contrastive Learning of Representations for Videos}
%
\author{Fanyi Xiao\thanks{\it Work done while at Amazon, now at Meta AI}\orcidlink{0000-0002-9839-1139} \and
Joseph Tighe$\,$\orcidlink{0000-0002-0716-8119} \and
Davide Modolo$\,$\orcidlink{0000-0002-7625-7748}}
%
\authorrunning{Fanyi Xiao et al.}
%
\institute{AWS AI Labs\\
\email{fyxiao@ucdavis.edu}, \email{\{tighej,dmodolo\}@amazon.com}}
\maketitle

\begin{abstract}
We present MaCLR, a novel method to explicitly perform cross-modal self-supervised video representations learning from visual and motion modalities.
Compared to previous video representation learning methods that mostly focus on learning motion cues implicitly from RGB inputs, MaCLR enriches standard contrastive learning objectives for RGB video clips with a cross-modal learning objective between a Motion pathway and a Visual pathway. 
We show that the representation learned with our MaCLR method focuses more on foreground motion regions and thus generalizes better to downstream tasks. 
To demonstrate this, we evaluate MaCLR on five datasets for both action recognition and action detection, and demonstrate state-of-the-art self-supervised performance on all datasets. 
Furthermore, we show that MaCLR representation can be as effective as representations learned with full supervision on UCF101 and HMDB51 action recognition, and even outperform the supervised representation for action recognition on VidSitu and SSv2, and action detection on AVA.  
\end{abstract}

\section{Introduction} \label{sec:intro}
Supervised learning has enjoyed great successes in many computer vision tasks in the past decade. One of the most important fuel in this successful journey is the availability of large amount of high-quality labeled data. Notably, the ImageNet~\cite{imagenet} dataset for image classification was the spark that ignited the deep learning revolution in vision.
In the video domain, the Kinetics dataset~\cite{kay-kinetics2017} has long been regarded as the ``ImageNet for videos'' and has enabled the ``pretrain-then-finetune'' paradigm for many video tasks.  
Interestingly, though years old, ImageNet and Kinetics are still the to-go datasets for pretraining that are publicly available. This shows how much effort is needed to create these large-scale labeled datasets. 

\begin{figure}[t]
    \centering
    \includegraphics[width=1.0\textwidth]{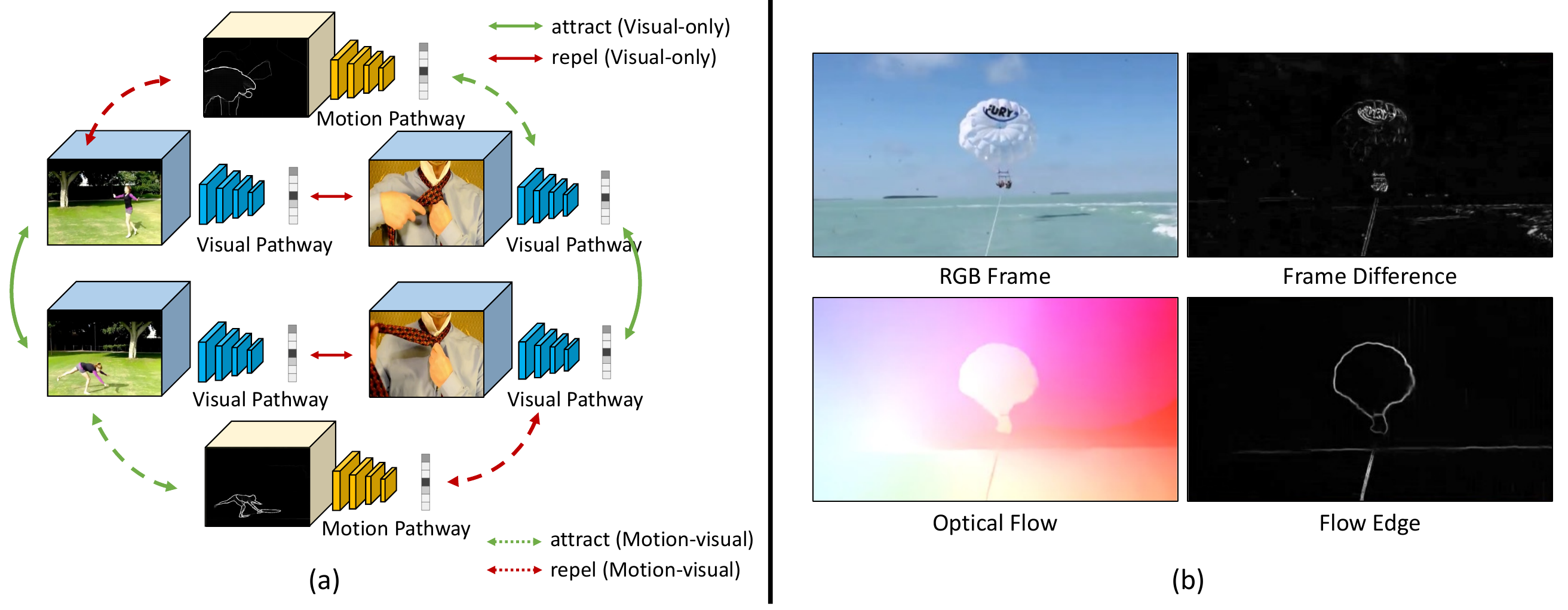}
    \caption{
    \textbf{(a)} \textbf{Cross-modal motion-visual learning}. We propose Motion-aware Contrastive Learning of Representations (\ours{}) as an explicit method to learn motion-aware video representations without labels -- Our visual pathway features are pushed to learn a representation aligns with our motion pathway and in doing so learn features that more robustly capture motion in the video. \textbf{(b)} \textbf{Motion inputs}. Given the RGB input (top-left), we compare three options for motion inputs. Best viewed on screen.
    }
    \label{fig:conceptandmotion}
\end{figure}

To mitigate the reliance on large-scale labeled datasets, \emph{self-supervised learning} came with the promise to learn useful representations from large amount of \emph{unlabeled} data. Following the recent success in NLP (\eg, BERT, GPT-3~\cite{devlin-bert2018,brown-gpt2020}), some works have attempted to find its counterpart in vision. Among them, pioneering research has been conducted in the image domain to produce successful methods like MoCo~\cite{he-moco2020} and SimCLR~\cite{chen-simclr2020}. Compared to images, large-scale video datasets induce even higher annotation costs, making it even more important to develop effective self-supervised methods to learn generalizable representations for videos.
Some recent video works attempted to learn such representations by training their models to solve pretext tasks, like predicting the correct temporal order of clips~\cite{misra-eccv2016,fernando-cvpr2017,brattoli-cvpr2017,wei-cvpr2018,buchler-eccv2018}, predict future frames~\cite{diba-dynamonet2019} and predict whether a video is played at its intrinsic speed~\cite{benaim-speednet2020}. Though successful to a certain extent, these methods do not explicitly make use of motion information derived from the temporal sequence, which has been shown to be important for supervised action recognition tasks~\cite{simonyan-iclr2015,feichtenhofer-cvpr2016,wang-eccv2016}.



In this paper, we propose \ours{}, a novel self-supervised video representation learning method that \emph{explicitly} models motion cues during training. 
\ours{} (Motion-aware Contrastive Learning of Representations) consists of two pathways: Visual and Motion. It uses both pathways during self-supervised pretraining, but only transfers the Visual to downstream tasks. 
When trained alone, the Visual pathway learns from RGB inputs using the contrastive InfoNCE objective, which mostly focuses on visual semantic information.  To help enriching the representation of Visual and make it motion-aware, we introduce a Motion pathway trained on motion inputs. We then connect Motion to Visual using a novel cross-modal contrastive objective that enables the Motion pathway to guide the learning of Visual towards relevant motion cues. As our experiments show, this formulation leads to rich video representations that capture both visual semantics and motion patterns. 


To evaluate \ours{}, we perform self-supervised pretraining on Kinetics-400 and transfer its representation to 5 video datasets for both action recognition (UCF101~\cite{soomro-ucf2012}, HMDB51~\cite{kuehne-hmdb2011}, Something-Something~\cite{ssv2}, VidSitu~\cite{sadhu-vidsitu2021}) and action detection (AVA~\cite{gu-ava2018}).  Without bells and whistle, \ours{} outperforms all previous video self-supervised methods on all datasets, under all evaluation settings. For example, \ours{} improves top-1 accuracy by 17\% and 16.9\% on UCF101 and HMD51, over previous SOTA trained on Kinetics-400. Furthermore, on Something-Something, VidSitu and AVA, \ours{} even outperforms its fully-supervised counterparts, demonstrating the strength of our approach.

\section{Related Work}

\noindent{\bf Self-supervised image representation learning. } 
The goal of self-supervised image representation learning is to learn useful representations from large collections of unlabeled images. Early work focused on designing different pretext tasks with the intent of inducing generalizable semantic representations~\cite{doersch-iccv2015,noroozi-jigsaw2016,noroozi-iccv2017,zhang-color2016}. Though producing promising results, these methods could not match the performance of fully-supervised trained representations~\cite{kolesnikov-cvpr2019}, as it is hard to prevent the network from utilizing shortcuts to solve pretext tasks (\eg, ``chromatic aberration'' in context prediction~\cite{doersch-iccv2015}). This changed when researchers re-visited the decade-old technique of contrastive learning~\cite{hadsell-cvpr2006,wu-instdisc2018}. Some of these recent work started to successfully produce results that were comparable to those of supervised learning on images~\cite{he-moco2020,chen-simclr2020,chen-mocov2,misra-pirl2020,grill-byol2020,chen-simsiam2020,caron-swav2020}. Though related, these work were designed to learn from static images and thus cannot utilize the rich temporal information contained in videos.  

\noindent{\bf Self-supervised video representation learning.} 
Videos present unique opportunities to extract self-supervision 
and the literature offers different directions. 
The first line of research focuses on designing video-specific \emph{pretext tasks}.
Besides the work mentioned earlier~\cite{misra-eccv2016,fernando-cvpr2017,wei-cvpr2018,diba-dynamonet2019,benaim-speednet2020}, others attempt to learn video representations by either tracking across frames patches~\cite{wang-iccv2015}, pixels~\cite{wang-cvpr2019}, colors~\cite{vondrick-color2018}, predicting temporal context for videos~\cite{recasens-brave2021,wang-lstcl2021}, or by enforcing consistency along videos semantics and play speeds~\cite{huang-iccv2021}.
A more recent line of work overcomes the need for pretext tasks by leveraging the \emph{contrastive learning} paradigm~\cite{qian-cvrl2020,feichtenhofer-cvpr2021}. Though successful to a certain extent, none of above methods \emph{explicitly} make use of the important motion cues derived from the video temporal sequence.
To better exploit such important information, \cite{wang-vidssl2019} applies a pretext task of regressing motion statistics,~\cite{han-coclr20,gavrilyuk-iccv2021} mine and cluster RGB images with similar motion cues, while~\cite{mahendran-accv2018,sayed-ssl2018,huang-self2021} exploit the correspondences between RGB and motion \emph{pixels}.  
\ours{} belongs to this recent class of works that aim at improving video representation using motion cues. However, it differs from previous works in the way it utilizes visual-motion correspondence in a cross-modal contrastive framework at a higher level than pixels, which yields a a method that is simpler, more robust and achieves considerably better results.

\noindent{\bf Motion in video tasks.} 
Motion information has been heavily studied for many video tasks. As a prominent motion representation, optical flow has been utilized in many video action classification methods, either in the form of classical hand-crafted spatiotemporal features~\cite{laptev-cvpr2008,dalal-eccv2006,wang-iccv2013}, or serve as input to deep CNN systems trained with supervised learning~\cite{feichtenhofer-nips2016,feichtenhofer-cvpr2016,wang-eccv2016}. In contrast, our method focuses on exploiting motion information in the context of self-supervised learning. 
Beyond video classification, motion has also been exploited in many other tasks like video object detection~\cite{zhu-iccv2017,kang-cvpr2017,feichtenhofer-iccv2017,xiao-eccv2018}, video frame prediction~\cite{sedaghat-arxiv2016,li-eccv2018}, video segmentation~\cite{tsai-cvpr2016,bao-cnnmrf2018,cheng-cvpr2018}, object tracking~\cite{henriques-pami2014,bertinetto-eccv2016,perazzi-cvpr2017}, and 3D reconstruction~\cite{ummenhofer-cvpr2017}.

\section{\ours{}}\label{sec:approach}
We design \ours{} as a two-branch network consisting of a Visual pathway and a Motion pathway (Fig.~\ref{fig:conceptandmotion}a). The Visual pathway takes as input visual\footnote{\it Sometimes also referred to as ``RGB'' in the literature.} clips and produces their visual embeddings. Similarly, the Motion pathway operates on motion clips (we will study different motion inputs in Sec.~\ref{sec:motioninputs}) and generates motion embeddings. 
\ours{} is trained using  three contrastive learning objectives (Sec.~\ref{sec:trainingmodist}): (i) a visual-only loss 
that pulls together visual clip embeddings that are sampled from the same video (solid green arrow in Fig.~\ref{fig:conceptandmotion}a) and pushes away that of different videos (solid red arrow); (ii) a motion-only loss that operates like (i), but on motion clips (omitted in Fig.~\ref{fig:conceptandmotion}a to avoid clutter) and 
(iii) a motion-visual loss to enforce alignment between embeddings of the visual and motion inputs (dashed arrows). 
As shown in Fig.~\ref{fig:conceptandmotion}a, we generate positive pairs from clips extracted from the same video (green arrows) and negative pairs from clips extracted from different videos (red arrows). 
After pretraining with \ours{}, \emph{we then remove the Motion pathway and transfer the Visual pathway to target datasets for task-specific finetuning}.

\subsection{Training \ours{}}\label{sec:trainingmodist}

\noindent{\bf Visual-only learning.} We model this using a contrastive learning objective. Similar to~\cite{qian-cvrl2020}, our model takes as input random clips with spatiotemporal jitterring. 
As shown in Fig.~\ref{fig:conceptandmotion}a, given a random clip we produce its embedding $v^q$ (query), and sample a second positive clip from the same video and produce its embedding $v^k$ (key), as well as $N$ negative embeddings $v^n_i$, $i\in\{1,...,N\}$ from other videos. Then, we train the Visual pathway with the InfoNCE objective $\mathcal{L}_{v}=\text{IN}(v^q, v^k, v^n)$~\cite{oord-arxiv2018,he-moco2020}: 
\begin{align}
\mathcal{L}_{v} = -\log \frac{\exp(v^q{\cdot}v^k / \tau)}{\exp(v^q{\cdot}v^k / \tau) + \sum_{i=1}^{N}\exp(v^q{\cdot}v^n_i  / \tau)}, 
\label{eq:lv}
\end{align}
where $\tau$ is a temperature parameter. This objective ensures that our Visual pathway pulls together embeddings $v^q$ and $v^k$, while pushing away those of all the negative clips $v^n_i$. \\

\noindent{\bf Motion-only learning.}
To improve the discriminativeness of the Motion pathway, we add another InfoNCE objective $\mathcal{L}_{m}=\text{IN}(m^q, m^k, m^n)$, which is trained in a similar way to $\mathcal{L}_{v}$ but this time on motion embeddings $m^q$, $m^k$ (both are sampled from the same video as $v_q$) and $m^n$ (which denotes a set of negative motion embeddings). This ensures that the Motion pathway is able to embed similar motion patterns close to each other. \\

\noindent{\bf Motion-Visual learning.}
We model this also with a contrastive learning objective, but with a different purpose compared to the previous two.
Here, we aim at enriching the Visual pathway to be motion-aware with the help of the Motion pathway.
Specifically, we train the model using the following InfoNCE objectives: 
\begin{align}\label{eq:lm2v}
\mathcal{L}_{mv} = \text{IN}(v^q, m^k, v^n) + \text{IN}(m^q, v^k, m^n).
\end{align}
Note that $v^q$ is \emph{not necessarily in temporal synchronization} with $m^k$, but rather just a motion clip sampled from the same video (same for $v^k$ and $m^q$). In our ablation, we show that allowing for this misalignment encourages the embedding to better learn semantic abstraction of visual and motion patterns, which leads to better performance.

One key difference to visual-only contrastive learning is on how we sample motion clips for both motion-only and motion-visual learning. Instead of sampling randomly, we constrain to only sample in temporal regions with strong motion cues. Specifically, we compute the sum of pixels $P_i$ on the motion input and only sample frames with $\sum_{i=1}^K P_i / K > \gamma$, where $K$ is the total number of pixels in a frame and $\gamma$ is the threshold. This process helps avoid sampling irrelevant regions with no motion and thus leads to better representations. \\

\noindent{\bf Final training objective.}
The final training objective for \ours{} is the sum of all aforementioned loss functions:
\begin{align}\label{eq:loss}
\mathcal{L} = \mathcal{L}_{v} + \mathcal{L}_{m} + \mathcal{L}_{mv}.
\end{align}
Training \ours{} end-to-end is non-trivial, as video representation are expensive to compute and to maintain (as contrastive learning requires large batch sizes~\cite{chen-simclr2020}). Inspired by~\cite{wu-instdisc2018,he-moco2020}, we solve this problem by adopting the idea of memory bank for negative samples. Specifically, we construct two memory banks of negative samples for visual and motion inputs, and maintain a momentum version of the Motion and Visual pathways updated as a moving average of their online counterparts with momentum coefficient $\lambda$: $\theta'~{\leftarrow}~\lambda\theta' + (1-\lambda)\theta$, where $\theta$ and $\theta'$ are weights for the online and momentum version of the model respectively. 
One caveat is that when pushing negatives into the pool, we push the video index, along with the embedding, so that we can avoid sampling visual or motion clips that are from the same video as positive clips, which would otherwise confuse the network and hurt the representations. 
Similar to~\cite{he-moco2020}, we forward queries through the online model and keys through the momentum model to produce embeddings.

\subsection{Design of motion inputs}\label{sec:motioninputs}
There are many ways to represent a motion input. A straightforward way is to directly compute the difference of pixel values between two consecutive frames.
While capturing motion to a certain extent, it also captures undesired signals like pixel value shifts caused by background motion (\eg, sea-wave in Fig.~\ref{fig:conceptandmotion}b top-right). A more appropriate representation might be optical flow~\cite{brox-pami2011,weinzaepfel-iccv2013,flownet,teed-raft2020}. However, a disadvantage of feeding in raw optical flow (or flow vector magnitude, as used in~\cite{fragkiadaki-cvpr2015}) is that it is heavily influenced by factors like illumination change (Fig.~\ref{fig:conceptandmotion}b bottom-left) and it also captures absolute flow magnitude, which is not very useful for learning general motion patterns. 
To overcome these limitations, in \ours{}, we propose to use flow edge maps as inputs to the Motion pathway network. Specifically, we apply a Sobel filter~\cite{sobel-sobel2014} onto the flow magnitude map to produce the flow edges (Fig.~\ref{fig:conceptandmotion}b bottom-right). In our experiments, this simple operation turns out to produce significantly better motion representations that focus on foreground motion regions.

\newcommand{\blocks}[3]{\multirow{3}{*}{\(\left[\begin{array}{c}\text{1$\times$1$^\text{2}$, #2}\\[-.1em] \text{1$\times$3$^\text{2}$, #2}\\[-.1em] \text{1$\times$1$^\text{2}$, #1}\end{array}\right]\)$\times$#3}
}
\newcommand{\blockt}[3]{\multirow{3}{*}{\(\left[\begin{array}{c}\text{{3$\times$1$^\text{2}$}, #2}\\[-.1em] \text{1$\times$3$^\text{2}$, #2}\\[-.1em] \text{1$\times$1$^\text{2}$, #1}\end{array}\right]\)$\times$#3}
}
\newcommand{\blocka}[3]{\multirow{3}{*}{\(\left[\begin{array}{c}\text{1$\times$1, #2}\\[-.1em] \text{3$\times$3, #2}\\[-.1em] \text{1$\times$1, #1}\end{array}\right]\)$\times$#3}
}
\newcommand{\blocktf}[3]{\multirow{3}{*}{\(\left[\begin{array}{c}\text{1$\times$1, #2}\\[-.1em] \text{[3$\times$1, 1$\times$3], #2}\\[-.1em] \text{1$\times$1, #1}\end{array}\right]\)$\times$#3}
}
\newcommand{\outsizes}[7]{\multirow{#7}{*}{\(\begin{array}{c} \text{\emph{Slow}}: \text{#1$\times$#2$^\text{2}$}\\[-.1em] \text{\emph{Fast}}: \text{#4$\times$#5$^\text{2}$}\end{array}\)}
}

\subsection{Visual and Motion pathway architectures}
\label{sec:arch}
Our \emph{Visual pathway} is a 3D ResNet50 (R3D-50) with a structure similar to that of ``Slow-only'' in~\cite{slowfast,qian-cvrl2020}, which features 2D convs in res$_2$, res$_3$ and non-degenerate 3D convs in res$_4$, res$_5$. It takes as input a tensor of size 3\x8\x224$^2$, capturing 8 frames of size 224$\times$224. The sampling stride is 8, which means that the visual input clip spans 8\x8 frames, corresponding to \app 2 seconds for videos at 30 FPS. To have larger temporal receptive field, we set the temporal kernel size of {\tt conv$_1$} to 5 following~\cite{qian-cvrl2020}. 

Our \emph{Motion pathway} is a 2D ResNet50. and it takes as input a tensor of size 3\x16\x224$^2$, stacking 16 motion frames. We use a sampling stride of 4, so that it spans for the same time as the visual input (i.e., \app 2 secs). 
Following the design philosophy of SlowFast Networks~\cite{slowfast}, we design our Motion pathway to be much more lightweight compared to our Visual pathway (1/8 channel sizes across the network), as motion inputs have intrinsically less variability (\ie, no variations on colors, illumination, \etc).  


\section{Experiments}

\subsection{Implementation Details} \label{sec:id}

\noindent{\bf \ours{} training details.} 
We train \ours{} on the Kinetics-400 (K400) dataset (CC-BY-4.0)~\cite{kay-kinetics2017}. The dataset consists of~\app240k video clips that span at most 10 seconds. These were originally annotated with 400 different action classes, but we \emph{do not} use any of these labels. We train \ours{} for 600 epochs on the whole 240k videos when we compare against the literature. For our ablation study, instead, we compare different variants of \ours{} trained for 100 epochs on a subset of 60k videos (``K400-mini''). 
We use a pool size ($N$ in Eq.~\ref{eq:lv}) of 65536 negative samples for both visual and motion inputs. We set the momentum update coefficient $\lambda=0.999$ and temperature $\tau$ to 0.1. The embedding dimension is set to 128 for both Visual and Motion pathways. 
For the visual inputs, we apply random spatial cropping, temporal jittering, $p$ = 0.2 probability grayscale conversion, $p$ = 0.5 horizontal flip, $p$ = 0.5 Gaussian blur, and $p$ = 0.8 color perturbation on brightness, contrast and saturation, all with 0.4 jittering ratio. For motion inputs, we randomly sample flow edge clips in high motion regions (with motion threshold $\gamma$ set to 0.02) and skip other augmentations. Our codebase is based on PySlowFast~\cite{fan-pyslowfast2020}.

\noindent{\bf Flow Edge Maps.}
To compute flow edge map for frame $t$, we first compute optical flow from frame $t$ to $t-5$, using {\tt RAFT-things}~\cite{teed-raft2020} model trained entirely on synthetic data without human annotations.
We hypothesize it would also work with flow computed from closer pairs, as long as the motion threshold $\gamma$ is adjusted accordingly.
Then, we apply a Sobel filter onto the magnitude map of optical flow and clamp the resulting edge map in [0, 10] as the final flow edge map. 
We note that this is an offline pre-processing that only needs to be done once and reused throughout training (and never during inference).

\noindent{\bf Baselines.} We compare against two baselines: (i) {\it Self-Supervised Visual-only} is a strong self-supervised representation trained from RGB inputs using only the contrastive learning objective of Eq.~\ref{eq:lv} (i.e., without our motion learning objectives $\mathcal{L}_{mv}$ and $\mathcal{L}_{m}$); and (ii) {\it Supervised}  is a fully supervised model trained for action classification on K400. Both baselines use a R3D-50 backbone.

\begin{table*}[t!]
  \centering 
  \captionsetup[subfloat]{captionskip=2pt}
  \captionsetup[subffloat]{justification=centering}
  \subfloat[\textbf{Motion-visual learning}\label{table:ablation-modist}]{
    \tablestyle{1pt}{1.05}
    \begin{tabular}{l | c | c c}
      \multicolumn{1}{c|}{method} & data & UCF & HMDB 
      \\
      \shline
      V-only & K400-mini & 63.6 & 33.7 
      \\
      M-only & K400-mini & 66.4 & 45.1 
      \\
      \textbf{\ours{}} & K400-mini & \textbf{78.1} & \textbf{47.2}
      \\
      \hline
      V-only & K400 & 74.6 & 46.3 
      \\
      \textbf{\ours{}} & K400 & \textbf{85.5} & \textbf{57.7}  
  \end{tabular}} \hfill
  \subfloat[\textbf{Motion inputs}\label{table:ablation-motioninputs}]{
    \tablestyle{0.5pt}{1.2}
    \begin{tabular}{l | c c}
      \multicolumn{1}{c|}{inputs} & UCF & HMDB
      \\
      \shline
      Diff & 71.6 & 40.1
      \\
      Flow & 74.1 & 44.2 
      \\
      \textbf{Edge} & \textbf{78.1} & \textbf{47.2}
  \end{tabular}}\hfill
  \subfloat[\textbf{Dissect components} \label{table:ablation-components}]{
    \tablestyle{0.5pt}{1.05}
    \begin{tabular}{l | c c}
      components & UCF & HMDB
      \\
      \shline
      \textbf{\ours{}} & \textbf{78.1} & \textbf{47.2}
      \\
      $-$t. jitter & 77.4 & 47.1
      \\
      $-$m. thresh & 77.3 & 46.4
      \\
      $-$$\mathcal{L}_{m}$ & 77.8 & 46.6
  \end{tabular}}
  \caption{
  \textbf{Ablating \ours{}}. 
  We present top-1 classification accuracy using the Linear Layer Training evaluation protocol (sec.~\ref{sec:ucf_hmdb}). In (a), V-only and M-only refers to the visual and motion only pretraining. In (b), Diff, Flow and Edge refer to motion inputs in the form of Frame Difference, Optical Flow and Flow Edges, respectively. Experiments in (b) and (c) are conducted on K400-mini. We use 8\x8 R3D-50 model for finetuning. 
  }
  \label{tab:ablation}
\end{table*}

\subsection{Action Recognition on UCF101 and HMDB51} \label{sec:ucf_hmdb}
\noindent{\bf Datasets and evaluation protocol.}
We first evaluate \ours{} for action recognition on the two most popular datasets in the literature: UCF101~\cite{soomro-ucf2012} and HMDB51~\cite{kuehne-hmdb2011} {(CC-BY-3.0)}. We follow the standard settings to perform self-supervised training on K400 and then transfer the learned weights to target datasets for evaluation. 
Two evaluation protocols are employed in the literature to evaluate the quality of the self-supervised representation: (i) {\it Linear Layer Training} freezes the trained backbone and simply trains a linear classifier on the target dataset, while (ii) {\it Full Network Training} finetunes the entire network end-to-end on the target dataset. For completeness, we evaluate using both protocols and report action classification top-1 accuracy.  
For all experiments on UCF101 and HMDB51, we report results using {\tt split1} for train/test split. In total, there are 9.5k/3.7k train/test videos with 101 action classes in UCF101, and 3.5k/1.5k train/test videos with 51 actions in HMDB51. We use the standard 10 (temporal) \x 3 (spatial) crop sampling during test~\cite{wang-nonlocal2018,slowfast}. We use these two datasets to compare against the state-of-the-art (SOTA). Additionally, we use K400-mini to conduct an extensive ablation study on the components of \ours{}. For the comparison with SOTA, we pretrain \ours{} with 8\x8 inputs for 600 epochs on K400, and finetune with 32\x8 inputs on downstream tasks, as these leads to the best performance. In our ablation study instead we simplify these settings for efficiency and pretrain for only 100 epochs and use 8\x8 inputs for finetuning. \\

\noindent{\bf Ablation: motion-visual learning (Table~\ref{table:ablation-modist}).}
First and foremost, we study the importance of enriching visual embeddings with motion cues using the proposed motion-visual learning objective of Eq.~\ref{eq:lm2v}. Results show that \ours{} improves substantially over Visual-only on both UCF and HMDB, when pretrained with either K400 or K400-mini. To understand if the benefit comes purely from the new motion objective $\mathcal{L}_{m}$, we  also trained a Motion-only model on K400-mini. Interestingly, this model performs slightly better than Visual-only, but much worse than \ours{}, showing the importance of training a video representation that can capture \emph{both} semantic and motion features. Finally, note how \ours{} trained on K400-mini also outperforms the Visual-only baseline pretrained on the full K400 ($4\times$ more data): +3.5/+0.9 on UCF/HMDB. \\

\noindent{\bf Ablation: motion representations (Table~\ref{table:ablation-motioninputs}).}
In Sec.~\ref{sec:approach} we discussed some conceptual advantages of using flow edge maps and here we evaluate it against two popular motion alternatives: Frame Difference and Optical Flow. As shown in Table~\ref{table:ablation-motioninputs}, Flow Edges is indeed the best way to represent motion for self-supervised training, thanks to its ability to prune background motion noise and absolute motion magnitude. That being said, even the much weaker Frame Difference representation outperforms the Visual-only baseline (Table~\ref{table:ablation-modist}) by +8.0 top-1 accuracy on UCF and +6.4 on HMDB. This further confirms the importance of enriching video representations with motion cues.\\

\noindent{\bf Ablation: \ours{} components (Table~\ref{table:ablation-components}).}
We now dissect \ours{} to study the importance of its components. \\
\noindent {\it Temporal Jittering.} Unlike previous work that learn self-supervised representation by exploiting pixel-level correspondences between RGB and optical flow inputs~\cite{mahendran-accv2018,sayed-ssl2018}, we demonstrate that it's more effective to learn self-supervised representations by introducing temporal ``misalignment'' between them. Specifically, we compare \ours{}, which trains on RGB and motion clips that are temporally jittered, against a variant that is trained on synchronized RGB and motion clips (i.e., sync pairs [$v^q$, $m^k$] and [$m^q$, $v^k$] in Eq.~\ref{eq:lm2v}). Our results show that the misaligned inputs lead to better representations (+0.7 on UCF), as it prevents the model from exploiting the shortcut of finding pixel correspondences using low-level visual cues. \\
\noindent {\it Motion thresholding.} Next, we study the motion input sampling strategy discussed in Sec.~\ref{sec:trainingmodist}. We compare \ours{} to a variant which randomly samples motion input clips, without removing those with little motion (i.e., setting threshold $\gamma=0$, Sec.~\ref{sec:id}). Without this threshold, top-1 accuracy degrades by -0.8 on both datasets, due to the noise introduced by clips with too little motion. \\
\noindent {\it Motion loss $\mathcal{L}_{m}$.} Finally, we study whether it's necessary to have the extra contrastive objective $\mathcal{L}_{m}$ between motion inputs (Eq.~\ref{eq:loss}), which is included to help training more discriminative motion embeddings. Results show that this motion discrimination objective is indeed useful as it improves top-1 acc by +0.3 and +0.6 on UCF101 and HMDB51.\\

\begin{table*}[!t]
\scriptsize
  \centering
    \begin{tabular}{ll|lcccccc} 
    Method            & Date                & Data (duration)    & Arch.       & Size        & Modality  & Frozen      & UCF         & HMDB        \\ 
    \shline
    MemDPC~\cite{han-memdpc2020}& 2020          & K400 (28d)          & R-2D3D-34     & 224$^2$       & V       & \cmark      & 54.1        & 30.5        \\
    MIL-NCE~\cite{miech-milnce2020}     & 2020          & HTM (15y)           & S3D         & 224$^2$       & V+T     & \cmark      & 82.7        & 53.1        \\
    MIL-NCE~\cite{miech-milnce2020}     & 2020          & HTM (15y)           & I3D         & 224$^2$       & V+T     & \cmark      & 83.4        & 54.8        \\
    XDC~\cite{alwassel-xdc2020}     & 2020          & IG65M~(21y)         & R(2+1)D       & 224$^2$       & V+A     & \cmark      & 85.3        & 56.0        \\ 
    ELO~\cite{piergiovanni-elo2020}   & 2020          & YT-8M (8y)        & R(2+1)D       & 224$^2$       & V+A     & \cmark      & --        & 64.5        \\ 
    AVSlowFast~\cite{xiao-avslowfast2019} & 2020          & K400 (28d)          & AVSlowFast-50 & 224$^2$     & V+A       & \cmark      & {77.4}      & {42.2}      \\    
    CoCLR~\cite{han-coclr20}                & 2020          & K400 (28d)          & S3D         & 128$^2$     & V       & \cmark      & {74.5}      & {46.1}      \\    
    CVRL~\cite{qian-cvrl2020}       & 2021            & K400 (28d)          & R3D-50      & 224$^2$     & V     & \cmark      & 89.8        & 58.3        \\ 
    MLFO~\cite{qian-iccv2021} & 2021            & K400 (28d)          & R3D-18      & 112$^2$     & V     & \cmark      &  63.2        &   33.4        \\ 
    BraVe~\cite{recasens-brave2021} & 2021            & K600 (36d)          & R3D-50      & 224$^2$     & V     & \cmark      &  88.8        &  61.8        \\ 
    \rowcolor{GrayBG}
    \textbf{\ours{}}      &             & K400 (28d)          & R3D-18      & 128$^2$       & V       & \cmark      & \bf{90.4}        & \bf{57.5}        \\
    \rowcolor{GrayBG}
    \textbf{\ours{}}      &             & K400 (28d)          & R3D-50      & 224$^2$       & V       & \cmark      & \bf{91.5}   & \bf{63.0}   \\
    \hline\hline
    \demph{w/o Pretrain}    &                   & \demph{-}  & \demph{R3D-50}  & \demph{224$^2$} & \demph{V} & \demph{\xmark}  & \demph{69.0}    & \demph{22.7}    \\ 
    CBT~\cite{sun-cbt2019}          & 2019            & K600+ (273d)        & S3D         & 112$^2$       & V     & \xmark      & 79.5        & 44.6        \\
    DynamoNet~\cite{diba-dynamonet2019}   & 2019            & YT-8M-1 (58d)       & STCNet      & 112$^2$       & V     & \xmark      & 88.1        & 59.9        \\ 
    XDC~\cite{alwassel-xdc2020}     & 2020            & IG65M~(21y)         & R(2+1)D       & 224$^2$       & V+A     & \xmark      & 94.2        & 67.4        \\ 
    AVSlowFast~\cite{xiao-avslowfast2019} & 2020          & K400 (28d)          & AVSlowFast-50 & 224$^2$     & V+A       & \xmark      & {87.0}      & {54.6}      \\    
    SpeedNet~\cite{benaim-speednet2020}   & 2020            & K400 (28d)          & S3D-G       & 224$^2$       & V     & \xmark      & 81.1        & 48.8        \\
    MemDPC~\cite{han-memdpc2020}  &2020           & K400 (28d)          & R-2D3D-34     & 224$^2$       & V     & \xmark      & 86.1        & 54.5        \\
    CoCLR~\cite{han-coclr20}  &  2020         & K400 (28d)          & S3D         & 128$^2$       & V       & \xmark      & {87.9}      & {54.6}        \\
    GDT~\cite{patrick-gdt2020}      & 2020            & K400 (28d)          & R(2+1)D       & 112$^2$       & V+A     & \xmark      & 89.3        & 60.0        \\ 
    GDT~\cite{patrick-gdt2020}      & 2020            & IG65M~(21y)         & R(2+1)D       & 112$^2$       & V+A     & \xmark      & 95.2        & 72.8        \\ 
    MIL-NCE~\cite{miech-milnce2020}   & 2020            & HTM (15y)           & S3D         & 224$^2$       & V+T     & \xmark      & 91.3        & 61.0        \\
    ELO~\cite{piergiovanni-elo2020}   & 2020          & YT-8M-2 (13y)       & R(2+1)D       & 224$^2$       & V+A     & \xmark      & 93.8        & 67.4        \\ 
    CVRL~\cite{qian-cvrl2020}       & 2021            & K400 (28d)          & R3D-50      & 224$^2$     & V     & \xmark      & 92.9        & 67.9        \\ 
    MLFO~\cite{qian-iccv2021} & 2021            & K400 (28d)          & R3D-18      & 112$^2$     & V     & \xmark      &  79.1        &  47.6        \\
    $\rho$BYOL~\cite{feichtenhofer-cvpr2021} & 2021            & K400 (28d)          & R3D-50      & 224$^2$     & V     & \xmark      &  \bf{94.2}        &  \bf{72.1}        \\
    MotionFit~\cite{gavrilyuk-iccv2021} & 2021            & K400 (28d)          & S3D-G      & 224$^2$     & V     & \xmark      &  90.1        &  50.6        \\
    ASCNet~\cite{huang-iccv2021} & 2021            & K400 (28d)          & S3D-G      & 224$^2$     & V     & \xmark      &  90.8        &  60.5        \\
    BraVe~\cite{recasens-brave2021} & 2021            & K600 (36d)          & R3D-50      & 224$^2$     & V     & \xmark      &  92.6        &  69.2        \\ 
    \rowcolor{GrayBG}
    \textbf{\ours{}}        &             & K400 (28d)          & R3D-18      & 128$^2$       & V       & \xmark      & 91.3        & 62.1        \\
    \rowcolor{GrayBG}
    \textbf{\ours{}}        &             & K400 (28d)          & R3D-50      & 224$^2$       & V       & \xmark      & 94.0      & 67.4        \\
    \rowcolor{GrayBG}
    \textbf{\ours{}}        &             & K400 (28d)          & R3D-50      & 224$^2$       & V+F     & \xmark      & \bf{94.2}      & 67.3        \\
    \hline
    \demph{Fully-Supervised~\cite{xie-eccv2018}}  &           & \demph{K400 (28d)}          & \demph{S3D}         & \demph{224$^2$}       & \demph{V}       & \demph{\xmark}      & \demph{96.8}        & \demph{75.9}      \\
    \end{tabular}
  \caption{
    \textbf{Comparison with state-of-the-art approaches.}
    We report top-1 accuracy.
    In parenthesis, we show the total video duration in time (\textbf{d} for day, \textbf{y} for year).
    The top half of the table contains results for the Linear protocol (Frozen \cmark), whereas the bottom half shows results for the Full end-to-end finetuning protocol (Frozen \xmark).  
    For Modality, V: visual only, A: audio, T: text narration.  
    }
  \label{table:sota-classification}
\end{table*}

\noindent{\bf Comparison to state-of-the-art (Table~\ref{table:sota-classification}).} 
We now compare \ours{} against previous self-supervised video representation learning methods in the literature using both the evaluation protocols introduced at the beginning of Sec.~\ref{sec:ucf_hmdb}: Linear (\cmark$\,$  for column ``Frozen'') and Full (\xmark).

By only training a linear layer on top of our self-supervised learned representation, our method is able to achieve significantly better top-1 classification accuracy compared to the previous state-of-the-art trained on K400: +16.7 and +16.9 over CoCLR on UCF101 and HMDB51, respectively. Only the recent CVRL method comes close to our results on UCF, but still lacks on HMDB ($-4.7$).
Moreover, \ours{} outperforms all previous methods, including those trained on 100\x more data than K400 (\eg, IG65M and Youtube8M), and those that use extra modalities like audio and text (\eg, XDC, MIL-NCE). 

Results using the end-to-end full training evaluation protocol show similar observations to the linear evaluation protocol: \ours{} again achieves competitive results among the methods trained on K400. 
When compared to previous approaches, only $\rho$BYOL, XDC and GDT produce results comparable to \ours{}. Among them, $\rho$BYOL is conceptually similar to our visual-only method, but augmented with the idea of sampling multiple clips ($\rho=4$) per video for training, which is complementary to our main contribution. On the other hand, both XDC and GDT are trained on 270\x more data (IG65M contains 21 years of video content \vs K400 only 28 days) and use extra audio modality as inputs. 
Furthermore, towards making the best effort in enabling fair comparison against the literature, we also present the results of a weaker \ours{} model trained with a smaller backbone (R18) and a smaller input size (128$\times$128). Under this setting, our model again convincingly outperforms models with similar backbone and input resolutions (\eg, 3D-RotNet, CBT, GDT, CoCLR).

We also tried to keep the Motion pathway during inference and ensemble its prediction with those of the Visual pathway (``V+F''). This produces results that are nearly identical to those obtained using only our motion-aware Visual pathway (``V''), which suggests that our novel training paradigm is indeed able to successfully ``distill'' motion information into the Visual pathway during pretraining. Finally, we also performed k-nearest-neighbor video retrieval to compare to the recent ASCNet work~\cite{huang-iccv2021} in Table~\ref{table:knn}. Despite similar accuracy when $k$=10, we largely outperform under the strictest 1-NN setting (+2.8\%), which shows the higher precision of our representations.

\noindent{\bf Low-shot finetuning.}
We further investigate how the performance of \ours{} varies with respect to the amount of data available for finetuning on the target task. We evaluate using the Full Training protocol on the UCF101 dataset starting from just 1\% of its training data (1 video per class) and gradually increase that to 100\% (9.5k videos). We compare results against our two baselines: Visual-only and Supervised (Table~\ref{table:lowshot}). \ours{} outperforms Visual-only across all training set sizes and it only requires 20\% of the training videos to match the performance of Visual-only with 100\% (89.1 vs 89.0). Another interesting observation is that the gap $\Delta$ between \ours{} and Visual-only reaches its maximum with the smallest training set (1\%), suggesting that motion-visual learning is particularly helpful for generalization in low-shot scenarios.

\begin{table*}[t!]
  \centering 
  \captionsetup[subfloat]{captionskip=2pt}
  \captionsetup[subffloat]{justification=centering}
  \subfloat[\textbf{kNN video retrieval}\label{table:knn}]{
    \tablestyle{1pt}{1.05}
    \begin{tabular}{l | c c c}
      \multicolumn{1}{c|}{method} & 1-NN & 10-NN
      \\
      \shline
      ASCNet R18 & 58.9 & 82.2 
      \\
      MaCLR R18 & \textbf{61.7} & \textbf{82.2} 
      \\
      \hline
      \demph{MaCLR R50} & \demph{73.4} & \demph{88.2} 
    \end{tabular}
    }  \hfill
  \subfloat[\textbf{Low-shot learning on UCF101}\label{table:lowshot}]{
    \tablestyle{1pt}{1.05}
    \begin{tabular}{l | c c c c c c c}
      \multicolumn{1}{c|}{method}  &    1\%     &  5\%  &  20\%      &  40\%      &  60\%      &  80\%      &  100\%  
      \\
      \shline
      \demph{Supervised}    &    \demph{69.3}   &   \demph{85.1}         &  \demph{93.0}  &  \demph{94.5}  &  \demph{94.7}  &  \demph{95.8}  &   \demph{95.4}
      \\
      \hline
      Visual-only  &   32.9    &   62.8  &   82.2  &   86.5  &   87.8  &   89.5  &   89.0
      \\
      \ours{}    &   42.8    &   71.9  &   89.1  &   91.3  &   92.9  &   93.4  &   94.0 
      \\
      {$\Delta$} &   +9.9    &   +9.1  &   +6.9  &   +4.8  &   +5.1  &   +3.9  &   +5.0
    \end{tabular}
    }
  \caption{
  \textbf{Video retrieval and low-shot learning on UCF101.} (a) reports kNN video retrieval results on split1 of UCF101. A video is considered to be correctly predicted if its ground-truth label is among the labels of its $k$ nearest neighbors retrieved from the training set.  (b) Rows indicate different pretrainings on K400, while columns vary the \% of UCF training data used for finetuning. All results are top-1 accuracy.
  }
  \label{tab:knnandlowshot}
\end{table*}

\subsection{Action Recognition on Something-Something}

Next we evaluate \ours{} on Something-Something-v2 (SSv2)~\cite{ssv2}, a challenging action classification dataset that is heavily focused on motion.  
Different from UCF101 and HMDB51 which contain action classes similar to K400, SSv2 contains a very different set of actions featuring complex human object interactions, like ``Moving something up'' and ``Pushing something from left to right''. The dataset consists of 168k training, 24k validation and 24k test videos, all annotated with 174 action classes. 
We finetune on SSv2 with a recipe that mostly follows the official implementation of~\cite{slowfast}: we use a clip size of 16\x8 and a batchsize of 16 (over 8 GPUs); we train for 22 epochs with an initial learning rate of 0.03 and decay it by 10\x~twice at 14 and 18 epochs; and a learning rate warm-up is scheduled for 0.19 epochs starting from a learning rate of 0.0001. 

We evaluate using both the Linear and Full finetune protocol.
We compare methods that are pretrained in different ways: \ours{} and the Visual-only baseline are pretrained self-supervisedly on K400, whereas R3D-50~\cite{slowfast} is pretrained with full supervision on K400. Rand Init is a randomly initialized network without pretraining (Table~\ref{table:ssv2}). 

For the Full protocol evaluation, it's clear that pretraining on K400 is beneficial and improves by almost +10 top-1 accuracy. Next, \ours{} outperforms the Visual-only baseline, showing once more the importance of learning from the added Motion pathway. Finally, when comparing to R3D-50 pretrained with full supervision, \ours{} not only closes the gap between self-supervised and fully-supervised methods, but even outperforms the supervised pretraining (+1.9). 

Furthermore, we test with the Linear protocol, which is much more challenging due to the large difference between the label spaces of K400 and SSv2.  
As expected, Table~\ref{table:ssv2} shows that the accuracy of all methods is much lower compared to their Full finetune results. 
However, it's notable that the gap between \ours{} and Visual-only significantly increases (+10.5 vs +2.5) compared to the Full protocol, which further demonstrates our method's generalization strength.  Moreover, it's interesting to see the supervised baseline underperform both self-supervised methods, as it's harder to overcome taxonomy bias under Linear protocol compared to the Full protocol for a representation pretrained with a fixed label taxonomy. 
We believe this is a promising example showing how self-supervised training can remove the label taxonomy bias that is inevitable under supervised settings, and lead to more general video representations that can be better transferred to new domains.  




\begin{table*}[t!]
  \centering 
  \captionsetup[subfloat]{captionskip=2pt}
  \captionsetup[subffloat]{justification=centering}
  \subfloat[\textbf{Action classification on SSv2}\label{table:ssv2}]{
    \tablestyle{1pt}{1.0}
    \begin{tabular}{l | c | c | c | c | c}
      \multicolumn{1}{c|}{method} & pretrain &  sup.  & acc@1 & acc@5 & rec@5 
      \\
      \shline
      \demph{Slow+NL} & \demph{K400} & \demph{\cmark} & \demph{29.1} & \demph{58.7} & \demph{\bf 19.2}
      \\
      \demph{R3D-50} & \demph{K400} & \demph{\cmark} & \demph{38.3} & \demph{69.3} & \demph{18.7}
      \\
      \hline
      Visual-only & K400 & \xmark & 32.8 & 61.6 & 13.6
      \\
      \rowcolor{GrayBG} \textbf{\ours{}} & K400 & \xmark & {\bf 43.0} & {\bf 73.2} & 17.5
    \end{tabular}
    } 
  \subfloat[\textbf{Verb prediction on VidSitu}\label{table:vidsitu}]{
    \tablestyle{1pt}{1.05}
    \begin{tabular}{l | c | c | c c}
      \multicolumn{1}{c|}{method} & pretrain &  sup.  &  Full &  Linear
      \\
      \shline
      \demph{R3D-50} & \demph{K400} & \demph{\cmark} &  \demph{55.5} &  \demph{16.3}
      \\
      \hline
      Rand Init & - & \xmark &  45.4 &  -
      \\
      Visual-only & K400 & \xmark &  54.9 &  16.6
      \\
      \rowcolor{GrayBG} \textbf{\ours{}} & K400 & \xmark &  {\bf 57.4} &  {\bf 27.1}
    \end{tabular}}
  \caption{
  \textbf{Results on SSv2 and VidSitu.} 
  (a) reports top-1 accuracy. For finetuning, we use 16\x8 clip as input following~\cite{slowfast}. 
  (b) reports top-1, top-5 accuracy and macro-averaged recall with five predictions on val set following~\cite{sadhu-vidsitu2021}. All models use a R3D-50 backbone with 16\x4 inputs. 
  }
\end{table*}

In this section we evaluate how \ours{} pretrained on YouTube-style short clips (K400) generalizes to a very different video domain: movie clips. For this, we evaluate our video representation on the recent VidSitu benchmark~\cite{sadhu-vidsitu2021} which features 30k movie clips from 3k different movies. Specifically, we benchmark on the \emph{verb prediction} task of VidSitu, which contains 1560 action classes (\eg, speak, walk, run, climb). We compare different pretraining strategies, using the same R3D-50 backbone with 16\x4 inputs, and we evaluate verb prediction results with top-1/top-5 accuracy and the macro-averaged recall metric with five predictions, as in~\cite{sadhu-vidsitu2021}. The results are shown in Table~\ref{table:vidsitu}. For the supervised ``R3D-50'' baseline, we pretrained its backbone using the K400 labels and then fine-tuned it on VidSitu. As for self-supervised pretraining, we evaluate both the Visual-only baseline and our \ours{}. Similar to our observations on SSv2, \ours{} outperforms all methods on both acc@1 and acc@5 metrics. The improvement over the Visual-only baseline is also particularly substantial, which suggests that motion information is particularly important to help  self-supervised representation generalize to different video domains.

\subsection{Action Detection on AVA}

In the previous section we showed that \ours{} can generalize to new video domains within the same downstream task (\ie, action recognition). However, we believe that our self-supervised representation can go beyond that and also generalize to novel downstream tasks, since it is not optimized for any task specific objective. To test this, we transfer \ours{} representation to the new task of action detection, which requires not only to recognize the action class, but also localize the person performing the action. 


\begin{table}[t]
  \centering
  \begin{tabular}{l | c | c | c}
    \multicolumn{1}{c|}{method} & pretrain dataset &  sup.  & mAP
    \\
    \shline
    \demph{Faster-RCNN~\cite{slowfast}} & \demph{ImageNet} & \demph{\cmark} & \demph{15.3}
    \\
    \demph{Faster-RCNN~\cite{slowfast}} & \demph{K400} & \demph{\cmark} & \demph{21.9}
    \\
    \hline
    Rand Init & - & \xmark & 6.6
    \\
    CVRL~\cite{qian-cvrl2020} & K400 & \xmark & 16.3
    \\
    Visual-only & K400 & \xmark & 18.6
    \\
    \rowcolor{GrayBG} \textbf{\ours{}} & K400 & \xmark & {\bf 22.1}
  \end{tabular}
  \caption{
  \textbf{Action detection on AVA.} We use 8\x8 clip as input for finetuning~\cite{slowfast}. CVRL numbers are taken from~\cite{qian-cvrl2020}. }
  \label{table:ava}
\end{table}

We evaluate action detection on the AVA dataset (CC-BY-4.0)~\cite{gu-ava2018} which contains 211k training and 57k validation videos. Spatiotemporal labels (i.e., action classes and bounding boxes) are provided at 1 FPS rate. We follow the standard evaluation protocol and compute mean Average Precision over 60 classes, using an IOU threshold of 0.5. 
We follow the Faster-RCNN detector design of \cite{slowfast} and use the Visual pathway architecture of Sec.~\ref{sec:arch} as the detector backbone. We fix the training schedule to 20 epochs with an initial learning rate of 0.1 and a batch size of 64~\cite{slowfast}. 

Results are shown in Table~\ref{table:ava}. Clearly, video pretraining plays a critical role in action detection, as demonstrated by the low mAP of 6.6 when training from scratch and the substantially lower AP achieved by supervised pretraining on ImageNet (pretrained 2D convs are inflated into 3D for fine-tuning~\cite{carreira-i3d2017}) compared to supervised pretraining on K400. As for self-supervised pretraining, both the Visual-only baseline and \ours{} outperform ImageNet supervised pretraining, again demonstrating the importance of pretraining on videos. Moreover, \ours{} again outperforms both the Visual-only baseline and the recent CVRL approach, which also only uses RGB inputs for pretraining. 

Finally, note how \ours{} even outperforms the supervised Faster-RCNN pretrained on K400. To the best of our knowledge, we are the first to demonstrate that self-supervised video learning can transfer to action detection and match the performance of fully-supervised pretraining. 


\begin{figure}[t]
    \centering
    \includegraphics[width=1.0\textwidth]{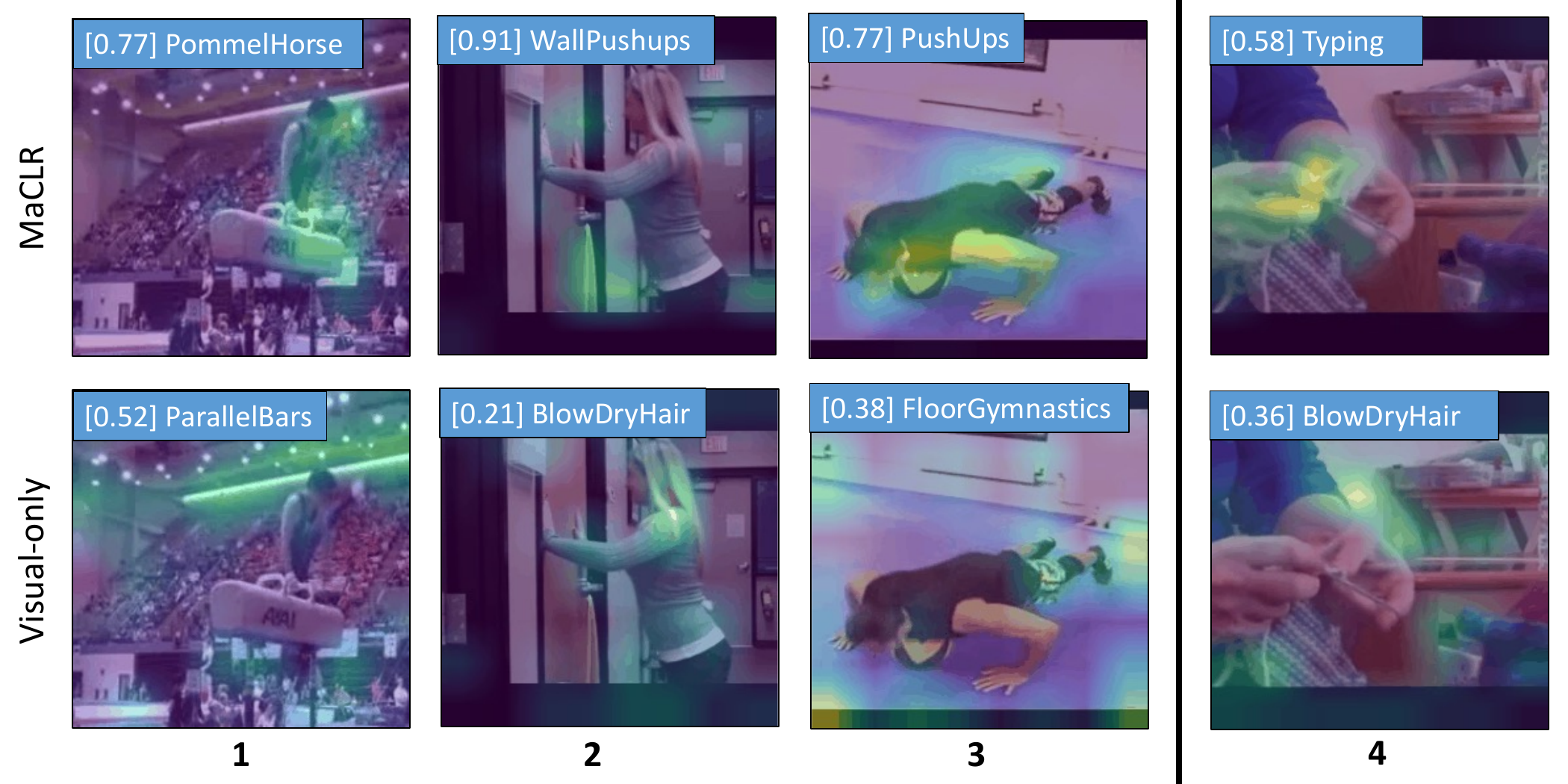}
    \caption{
    \textbf{Grad-CAM visualization for \ours{} (top) and Visual-only (bottom) representations}. Predictions are overlaid on each frame. 
    }
    \label{fig:gradcam}
\end{figure}

\subsection{Visualizing \ours{} Representations}\label{sec:gradcam}

To gain deeper insights on what \ours{} has learned in its representations, we adopt Grad-CAM~\cite{selvaraju-gradcam2017} to visualize the spatiotemporal regions that contribute the most to the classification decisions on UCF101. 
%
As shown in Fig.~\ref{fig:gradcam}, we observe that the representation learned by \ours{} focuses more on the ``motion-sensitive'' regions (\ie, regions where object motion likely occur). 
For example, in col-1, \ours{} makes the correct prediction of ``PommelHorse'' by focusing its attention on the person carrying out the motion. The Visual-only model, on the other hand, incorrectly predicted ``ParallelBars'' as it finds ``bar-like'' straight lines in the background. This pattern can also be observed in col-2 (Visual-only model predicts ``BlowDryHair'' after finding hair textures). 
Furthermore, we can observe another type of behavior in col-3. 
In both examples, the background scenes (gym) are associated with many fine-grained action classes (different gym activities), 
our model is able to distinguish them by focusing on the actual motion pattern. The baseline, instead, gets confused as it focuses too much on the background. 
Finally, we present a failure case in the last column where \ours{} correctly focuses on the right motion region (fingers), but confuses the finger motion of ``Knitting'' with ``Typing''.


\section*{Conclusion}
We presented \ours{} to learn self-supervised video representations with explicit cross-modal motion-visual contrastive learning.
We demonstrated SOTA self-supervised performance with \ours{} across various datasets and tasks. 
Moreover, we showed that \ours{} representations can be as effective as representations learned with full supervision for SSv2 action recognition, VidSitu verb prediction and AVA action detection. 
Given the simplicity of our method, we hope it will serve as a strong baseline for future research in self-supervised video representation learning.

%
%
\bibliographystyle{splncs04}
\bibliography{bibs}

\begin{thebibliography}{10}
\providecommand{\url}[1]{\texttt{#1}}
\providecommand{\urlprefix}{URL }
\providecommand{\doi}[1]{https://doi.org/#1}

\bibitem{ssv2}
{20BN-Something-Something Dataset V2}

\bibitem{alwassel-xdc2020}
Alwassel, H., Mahajan, D., Korbar, B., Torresani, L., Ghanem, B., Tran, D.:
  Self-supervised learning by cross-modal audio-video clustering. In: NeurIPS
  (2020)

\bibitem{bao-cnnmrf2018}
Bao, L., Wu, B., Liu, W.: {CNN in MRF: Video object segmentation via inference
  in a CNN-based higher-order spatio-temporal MRF}. In: CVPR (2018)

\bibitem{benaim-speednet2020}
Benaim, S., Ephrat, A., Lang, O., Mosseri, I., Freeman, W.T., Rubinstein, M.,
  Irani, M., Dekel, T.: {SpeedNet: Learning the speediness in videos}. In: CVPR
  (2020)

\bibitem{bertinetto-eccv2016}
Bertinetto, L., Valmadre, J., Henriques, J.F., Vedaldi, A., Torr, P.H.:
  Fully-convolutional siamese networks for object tracking. In: ECCV (2016)

\bibitem{brattoli-cvpr2017}
Brattoli, B., Buchler, U., Wahl, A.S., Schwab, M.E., Ommer, B.: {LSTM
  self-supervision for detailed behavior analysis}. In: CVPR (2017)

\bibitem{brown-gpt2020}
Brown, T.B., Mann, B., Ryder, N., Subbiah, M., Kaplan, J., Dhariwal, P.,
  Neelakantan, A., Shyam, P., Sastry, G., Askell, A., et~al.: Language models
  are few-shot learners. In: NeurIPS (2020)

\bibitem{brox-pami2011}
Brox, T., Malik, J.: {Large displacement optical flow: descriptor matching in
  variational motion estimation}. T-PAMI  (2011)

\bibitem{buchler-eccv2018}
Buchler, U., Brattoli, B., Ommer, B.: Improving spatiotemporal self-supervision
  by deep reinforcement learning. In: ECCV (2018)

\bibitem{caron-swav2020}
Caron, M., Misra, I., Mairal, J., Goyal, P., Bojanowski, P., Joulin, A.:
  Unsupervised learning of visual features by contrasting cluster assignments.
  In: NeurIPS (2020)

\bibitem{carreira-i3d2017}
Carreira, J., Zisserman, A.: Quo vadis, action recognition? a new model and the
  kinetics dataset. In: CVPR (2017)

\bibitem{chen-simclr2020}
Chen, T., Kornblith, S., Norouzi, M., Hinton, G.: A simple framework for
  contrastive learning of visual representations. In: ICML (2020)

\bibitem{chen-mocov2}
Chen, X., Fan, H., Girshick, R., He, K.: Improved baselines with momentum
  contrastive learning. arXiv preprint arXiv:2003.04297  (2020)

\bibitem{chen-simsiam2020}
Chen, X., He, K.: Exploring simple siamese representation learning. In: CVPR
  (2021)

\bibitem{cheng-cvpr2018}
Cheng, J., Tsai, Y.H., Hung, W.C., Wang, S., Yang, M.H.: Fast and accurate
  online video object segmentation via tracking parts. In: CVPR (2018)

\bibitem{dalal-eccv2006}
Dalal, N., Triggs, B., Schmid, C.: Human detection using oriented histograms of
  flow and appearance. In: ECCV (2006)

\bibitem{imagenet}
Deng, J., Dong, W., Socher, R., Li, L.J., Li, K., Fei-Fei, L.: {Imagenet: A
  Large-Scale Hierarchical Image Database}. In: CVPR (2009)

\bibitem{devlin-bert2018}
Devlin, J., Chang, M.W., Lee, K., Toutanova, K.: {BERT: Pre-training of deep
  bidirectional transformers for language understanding}. In: NAACL (2019)

\bibitem{diba-dynamonet2019}
Diba, A., Sharma, V., Gool, L.V., Stiefelhagen, R.: {DynamoNet: Dynamic action
  and motion network}. In: ICCV (2019)

\bibitem{doersch-iccv2015}
Doersch, C., Gupta, A., Efros, A.A.: {Unsupervised Visual Representation
  Learning by Context Prediction}. In: ICCV (2015)

\bibitem{flownet}
Dosovitskiy, A., Fischer, P., Ilg, E., Hausser, P., Hazrba, C., Golkov, V.,
  Smagt, P., Cremers, D., Brox, T.: Flownet: Learning optical flow with
  convolutional networks. In: ICCV (2015)

\bibitem{fan-pyslowfast2020}
Fan, H., Li, Y., Xiong, B., Lo, W.Y., Feichtenhofer, C.: Pyslowfast.
  \url{https://github.com/facebookresearch/slowfast} (2020)

\bibitem{slowfast}
Feichtenhofer, C., Fan, H., Malik, J., He, K.: {SlowFast Networks for Video
  Recognition}. In: ICCV (2019)

\bibitem{feichtenhofer-cvpr2021}
Feichtenhofer, C., Fan, H., Xiong, B., Girshick, R., He, K.: A large-scale
  study on unsupervised spatiotemporal representation learning. In: CVPR (2021)

\bibitem{feichtenhofer-nips2016}
Feichtenhofer, C., Pinz, A., Wildes, R.: Spatiotemporal residual networks for
  video action recognition. In: NeurIPS (2016)

\bibitem{feichtenhofer-cvpr2016}
Feichtenhofer, C., Pinz, A., Zisserman, A.: Convolutional two-stream network
  fusion for video action recognition. In: CVPR (2016)

\bibitem{feichtenhofer-iccv2017}
Feichtenhofer, C., Pinz, A., Zisserman, A.: Detect to track and track to
  detect. In: ICCV (2017)

\bibitem{fernando-cvpr2017}
Fernando, B., Bilen, H., Gavves, E., Gould, S.: Self-supervised video
  representation learning with odd-one-out networks. In: CVPR (2017)

\bibitem{fragkiadaki-cvpr2015}
Fragkiadaki, K., Arbelaez, P., Felsen, P., Malik, J.: Learning to segment
  moving objects in videos. In: CVPR (2015)

\bibitem{gavrilyuk-iccv2021}
Gavrilyuk, K., Jain, M., Karmanov, I., Snoek, C.G.: Motion-augmented
  self-training for video recognition at smaller scale. In: Proceedings of the
  IEEE/CVF International Conference on Computer Vision (2021)

\bibitem{grill-byol2020}
Grill, J.B., Strub, F., Altch{\'e}, F., Tallec, C., Richemond, P.H.,
  Buchatskaya, E., Doersch, C., Pires, B.A., Guo, Z.D., Azar, M.G., et~al.:
  Bootstrap your own latent: A new approach to self-supervised learning. In:
  NeurIPS (2020)

\bibitem{gu-ava2018}
Gu, C., Sun, C., Ross, D.A., Vondrick, C., Pantofaru, C., Li, Y.,
  Vijayanarasimhan, S., Toderici, G., Ricco, S., Sukthankar, R., et~al.: {AVA:
  A video dataset of spatio-temporally localized atomic visual actions}. In:
  CVPR (2018)

\bibitem{hadsell-cvpr2006}
Hadsell, R., Chopra, S., LeCun, Y.: Dimensionality reduction by learning an
  invariant mapping. In: CVPR (2006)

\bibitem{han-memdpc2020}
Han, T., Xie, W., Zisserman, A.: Memory-augmented dense predictive coding for
  video representation learning. In: ECCV (2020)

\bibitem{han-coclr20}
Han, T., Xie, W., Zisserman, A.: Self-supervised co-training for video
  representation learning. In: NeurIPS (2020)

\bibitem{he-moco2020}
He, K., Fan, H., Wu, Y., Xie, S., Girshick, R.: Momentum contrast for
  unsupervised visual representation learning. In: CVPR (2020)

\bibitem{henriques-pami2014}
Henriques, J.F., Caseiro, R., Martins, P., Batista, J.: High-speed tracking
  with kernelized correlation filters. T-PAMI  (2014)

\bibitem{huang-iccv2021}
Huang, D., Wu, W., Hu, W., Liu, X., He, D., Wu, Z., Wu, X., Tan, M., Ding, E.:
  {ASCNet: Self-supervised video representation learning with appearance-speed
  consistency}. In: ICCV (2021)

\bibitem{huang-self2021}
Huang, L., Liu, Y., Wang, B., Pan, P., Xu, Y., Jin, R.: Self-supervised video
  representation learning by context and motion decoupling. In: CVPR (2021)

\bibitem{kang-cvpr2017}
Kang, K., Li, H., Xiao, T., Ouyang, W., Yan, J., Liu, X., Wang, X.: Object
  detection in videos with tubelet proposal networks. In: CVPR (2017)

\bibitem{kay-kinetics2017}
Kay, W., Carreira, J., Simonyan, K., Zhang, B., Hillier, C., Vijayanarasimhan,
  S., Viola, F., Green, T., Back, T., Natsev, P., et~al.: The kinetics human
  action video dataset. arXiv preprint arXiv:1705.06950  (2017)

\bibitem{kolesnikov-cvpr2019}
Kolesnikov, A., Zhai, X., Beyer, L.: Revisiting self-supervised visual
  representation learning. In: CVPR (2019)

\bibitem{kuehne-hmdb2011}
Kuehne, H., Jhuang, H., Garrote, E., Poggio, T., Serre, T.: {HMDB: a large
  video database for human motion recognition}. In: ICCV (2011)

\bibitem{laptev-cvpr2008}
Laptev, I., Marszalek, M., Schmid, C., Rozenfeld, B.: Learning realistic human
  actions from movies. In: CVPR (2008)

\bibitem{li-eccv2018}
Li, Y., Fang, C., Yang, J., Wang, Z., Lu, X., Yang, M.H.: Flow-grounded
  spatial-temporal video prediction from still images. In: ECCV (2018)

\bibitem{mahendran-accv2018}
Mahendran, A., Thewlis, J., Vedaldi, A.: Cross pixel optical-flow similarity
  for self-supervised learning. In: ACCV (2018)

\bibitem{miech-milnce2020}
Miech, A., Alayrac, J.B., Smaira, L., Laptev, I., Sivic, J., Zisserman, A.:
  End-to-end learning of visual representations from uncurated instructional
  videos. In: CVPR (2020)

\bibitem{misra-pirl2020}
Misra, I., Maaten, L.v.d.: Self-supervised learning of pretext-invariant
  representations. In: CVPR (2020)

\bibitem{misra-eccv2016}
Misra, I., Zitnick, C.L., Hebert, M.: Shuffle and learn: unsupervised learning
  using temporal order verification. In: ECCV (2016)

\bibitem{noroozi-jigsaw2016}
Noroozi, M., Favaro, P.: Unsupervised learning of visual representations by
  solving jigsaw puzzles. In: ECCV (2016)

\bibitem{noroozi-iccv2017}
Noroozi, M., Pirsiavash, H., Favaro, P.: Representation learning by learning to
  count. In: ICCV (2017)

\bibitem{oord-arxiv2018}
Oord, A.v.d., Li, Y., Vinyals, O.: Representation learning with contrastive
  predictive coding. arXiv preprint arXiv:1807.03748  (2018)

\bibitem{patrick-gdt2020}
Patrick, M., Asano, Y.M., Kuznetsova, P., Fong, R., Henriques, J.F., Zweig, G.,
  Vedaldi, A.: Multi-modal self-supervision from generalized data
  transformations. In: ICCV (2021)

\bibitem{perazzi-cvpr2017}
Perazzi, F., Khoreva, A., Benenson, R., Schiele, B., Sorkine-Hornung, A.:
  Learning video object segmentation from static images. In: CVPR (2017)

\bibitem{piergiovanni-elo2020}
Piergiovanni, A., Angelova, A., Ryoo, M.S.: Evolving losses for unsupervised
  video representation learning. In: CVPR (2020)

\bibitem{qian-iccv2021}
Qian, R., Li, Y., Liu, H., See, J., Ding, S., Liu, X., Li, D., Lin, W.:
  Enhancing self-supervised video representation learning via multi-level
  feature optimization. In: ICCV (2021)

\bibitem{qian-cvrl2020}
Qian, R., Meng, T., Gong, B., Yang, M.H., Wang, H., Belongie, S., Cui, Y.:
  Spatiotemporal contrastive video representation learning. In: CVPR (2021)

\bibitem{recasens-brave2021}
Recasens, A., Luc, P., Alayrac, J.B., Wang, L., Strub, F., Tallec, C.,
  Malinowski, M., Patraucean, V., Altch{\'e}, F., Valko, M., et~al.: Broaden
  your views for self-supervised video learning. In: ICCV (2021)

\bibitem{sadhu-vidsitu2021}
Sadhu, A., Gupta, T., Yatskar, M., Nevatia, R., Kembhavi, A.: Visual semantic
  role labeling for video understanding. In: CVPR (2021)

\bibitem{sayed-ssl2018}
Sayed, N., Brattoli, B., Ommer, B.: Cross and learn: Cross-modal
  self-supervision. In: German Conference on Pattern Recognition (2018)

\bibitem{sedaghat-arxiv2016}
Sedaghat, N., Zolfaghari, M., Brox, T.: Hybrid learning of optical flow and
  next frame prediction to boost optical flow in the wild. arXiv preprint
  arXiv:1612.03777  (2016)

\bibitem{selvaraju-gradcam2017}
Selvaraju, R.R., Cogswell, M., Das, A., Vedantam, R., Parikh, D., Batra, D.:
  {Grad-CAM: Visual explanations from deep networks via gradient-based
  localization}. In: ICCV (2017)

\bibitem{simonyan-iclr2015}
Simonyan, K., Zisserman, A.: {Very Deep Convolutional Networks for Large-Scale
  Image Recognition}. In: ICLR (2015)

\bibitem{sobel-sobel2014}
Sobel, I.: History and definition of the sobel operator  (2014)

\bibitem{soomro-ucf2012}
Soomro, K., Zamir, A.R., Shah, M.: A dataset of 101 human action classes from
  videos in the wild. In: ICCV Workshops (2013)

\bibitem{sun-cbt2019}
Sun, C., Baradel, F., Murphy, K., Schmid, C.: Contrastive bidirectional
  transformer for temporal representation learning. arXiv preprint
  arXiv:1906.05743  (2019)

\bibitem{teed-raft2020}
Teed, Z., Deng, J.: Raft: Recurrent all-pairs field transforms for optical
  flow. In: ECCV (2020)

\bibitem{tsai-cvpr2016}
Tsai, Y.H., Yang, M.H., Black, M.J.: Video segmentation via object flow. In:
  CVPR (2016)

\bibitem{ummenhofer-cvpr2017}
Ummenhofer, B., Zhou, H., Uhrig, J., Mayer, N., Ilg, E., Dosovitskiy, A., Brox,
  T.: Demon: Depth and motion network for learning monocular stereo. In: CVPR
  (2017)

\bibitem{vondrick-color2018}
Vondrick, C., Shrivastava, A., Fathi, A., Guadarrama, S., Murphy, K.: Tracking
  emerges by colorizing videos. In: ECCV (2018)

\bibitem{wang-iccv2013}
Wang, H., Schmid, C.: Action recognition with improved trajectories. In: ICCV
  (2013)

\bibitem{wang-vidssl2019}
Wang, J., Jiao, J., Bao, L., He, S., Liu, Y., Liu, W.: Self-supervised
  spatio-temporal representation learning for videos by predicting motion and
  appearance statistics. In: CVPR (2019)

\bibitem{wang-lstcl2021}
Wang, J., Bertasius, G., Tran, D., Torresani, L.: Long-short temporal
  contrastive learning of video transformers. arXiv preprint arXiv:2106.09212
  (2021)

\bibitem{wang-eccv2016}
Wang, L., Xiong, Y., Wang, Z., Qiao, Y., Lin, D., Tang, X., {Van Gool}, L.:
  Temporal segment networks: Towards good practices for deep action
  recognition. In: ECCV (2016)

\bibitem{wang-nonlocal2018}
Wang, X., Girshick, R., Gupta, A., He, K.: Non-local neural networks. In: CVPR
  (2018)

\bibitem{wang-iccv2015}
Wang, X., Gupta, A.: {Unsupervised Learning of Visual Representations using
  Videos}. In: ICCV (2015)

\bibitem{wang-cvpr2019}
Wang, X., Jabri, A., Efros, A.A.: Learning correspondence from the
  cycle-consistency of time. In: CVPR (2019)

\bibitem{wei-cvpr2018}
Wei, D., Lim, J.J., Zisserman, A., Freeman, W.T.: Learning and using the arrow
  of time. In: CVPR (2018)

\bibitem{weinzaepfel-iccv2013}
Weinzaepfel, P., Revaud, J., Harchaoui, Z., Schmid, C.: Deepflow: Large
  displacement optical flow with deep matching. In: ICCV (2013)

\bibitem{wu-instdisc2018}
Wu, Z., Xiong, Y., Yu, S.X., Lin, D.: Unsupervised feature learning via
  non-parametric instance discrimination. In: CVPR (2018)

\bibitem{xiao-eccv2018}
Xiao, F., Lee, Y.J.: Video object detection with an aligned spatial-temporal
  memory. In: ECCV (2018)

\bibitem{xiao-avslowfast2019}
Xiao, F., Lee, Y.J., Grauman, K., Malik, J., Feichtenhofer, C.: Audiovisual
  slowfast networks for video recognition. arXiv preprint arXiv:2001.08740
  (2019)

\bibitem{xie-eccv2018}
Xie, S., Sun, C., Huang, J., Tu, Z., Murphy, K.: Rethinking spatiotemporal
  feature learning: Speed-accuracy trade-offs in video classification. In: ECCV
  (2018)

\bibitem{zhang-color2016}
Zhang, R., Isola, P., Efros, A.A.: Colorful image colorization. In: ECCV (2016)

\bibitem{zhu-iccv2017}
Zhu, X., Wang, Y., Dai, J., Yuan, L., Wei, Y.: Flow-guided feature aggregation
  for video object detection. ICCV  (2017)

\end{thebibliography}
\end{document}